\documentclass[letterpaper, 10 pt, conference]{ieeeconf}  

\IEEEoverridecommandlockouts                              

\usepackage{graphics} 
\usepackage{epsfig} 
\usepackage{mathptmx} 
\usepackage{times} 
\usepackage{amsmath} 
\usepackage{amssymb}  

\usepackage{array}
\usepackage{booktabs}
\usepackage{multirow}
\usepackage{siunitx}

\makeatletter
\let\NAT@parse\undefined
\makeatother

\usepackage{hyperref}

\title{\LARGE \bf
Navigating the Proximity-Safety Balance: Constraint Decomposition for Human Following in Pedestrian Crowds
}

\author{Shiting Gong$^{1*}$, Jianpeng Yao$^{2*}$, Jinfeng Wang$^{2}$, Marco Pavone$^{3,4}$ and Jiachen Li$^{5\dagger}$
\thanks{$^{*}$Equal Contribution}%
\thanks{$^{\dagger}$Corresponding author: {\tt\small jiachen\_li@gatech.edu}}%
\thanks{$^{1}$University of Pennsylvania, PA, USA}%
\thanks{$^{2}$University of California, Riverside, CA, USA}%
\thanks{$^{3}$Stanford University, CA, USA $^{4}$NVIDIA Research, CA, USA}%
\thanks{$^{5}$Georgia Institute of Technology, GA, USA}%
}

\begin{document}

\maketitle
\thispagestyle{empty}
\pagestyle{empty}

\begin{abstract}
Following a target human in crowded environments involves an inherent conflict between staying close to the target and navigating safely among surrounding pedestrians and obstacles. This conflict becomes more severe in dense scenarios, where aggressive following risks collisions and conservative margins lead to target loss, especially when pedestrian behaviors are unfamiliar or unpredictable. Existing reinforcement learning (RL) methods typically encode these competing objectives into a single dense reward, but the resulting proximity-safety balance is implicit and difficult to adjust across conditions. To address this, we decompose the human-following task into a sparse task reward and independent cost constraints within a multi-constraint RL formulation, where each constraint is managed through cost thresholds with direct behavioral meaning rather than implicit reward weight ratios, allowing explicit and tunable control over the trade-off. We further quantify the prediction uncertainty of human motions and integrate these estimates into the RL costs to enhance safety under unpredictable conditions. Extensive experiments across both in-distribution and out-of-distribution settings demonstrate that our method achieves an effective proximity-safety balance compared to baselines. Real-robot deployment further validates the feasibility of our method in real-world scenarios. 
More details are available on our project page: \url{https://nav-ps-balance.github.io/}.

\end{abstract}

\section{Introduction}

Robots are increasingly expected to follow specific individuals in applications such as healthcare, companionship, and assistance~\cite{li2023exploring}, requiring them to navigate safely among pedestrians and obstacles while maintaining close proximity to the target. This task involves an inherent conflict between proximity and safety.
Unlike point-goal navigation, where the robot can freely reroute around obstacles, human following involves a moving target, and any detour to avoid collisions risks increasing the distance to or losing the target entirely. Dense crowds further intensify this conflict, as safe passages shrink and the robot must frequently decide how to balance proximity against safety.

Learning-based methods offer advantages such as millisecond-level action generation via single forward inference and strong performance in in-distribution settings, making them attractive for real-time robot navigation. However, existing RL approaches typically combine all objectives into a single dense reward through weighted summation~\cite{kastner2022human, leisiazar2023mcts}. This formulation entangles the proximity-safety balance within the reward weights, where the trade-off is implicitly determined by their relative magnitudes, which interact in complex ways during training, cannot be independently adjusted for each behavioral aspect, and offer no guarantee that the resulting behavior reflects the designer's intended balance. This problem is exacerbated in real-world deployment, where pedestrian behaviors are diverse and hard to anticipate, and a fixed implicit trade-off learned in training may fail to generalize to these challenging conditions.

To address these limitations, we decompose the human-following task into a sparse task reward for target proximity and three independent cost constraints for following distance, human safety, and obstacle safety. Each constraint is monitored by a dedicated critic, and the trade-off is managed through cost thresholds with direct behavioral meaning, such as tolerable intrusions or a desired following distance, rather than reward weight ratios whose effect on policy behavior is implicit and cannot be directly verified.
To further enhance safety in unpredictable pedestrian environments, we quantify prediction uncertainty of human motions~\cite{gibbs2021adaptive, gibbs2024conformal, lindemann2023safe} and incorporate these estimates into both the observation space and cost formulation, allowing the policy to adopt more conservative behavior when confidence in its predictions is low. Our policy network combines an attention-based Transformer for social interaction modeling with a convolutional neural network for spatial constraint reasoning in a unified architecture~\cite{wang2024navformer}. The entire system is jointly optimized under a constrained reinforcement learning (CRL) framework using PPO-Lagrangian~\cite{ji2024omnisafe, ray2019benchmarking}.
To validate our approach, we extend the CrowdNav~\cite{chen2019crowd} simulator with static obstacles for evaluation under more realistic conditions. Across both in-distribution and out-of-distribution (OOD) scenarios with varying crowd densities, pedestrian behaviors, and environment layouts, our method achieves higher success rates and lower collision rates than optimization-based baselines and ablation variants. We further deploy the policy on a real robot under ROS 2 to validate its effectiveness in the real world.

The main contributions of this paper are as follows:
\begin{itemize}
\item We propose a novel framework that decomposes human-following objectives into a sparse task reward and independent cost constraints, with the trade-off managed through behaviorally meaningful cost thresholds and jointly optimized using PPO-Lagrangian.
\item We integrate prediction uncertainty of human motions into the observation space and cost formulation, and design a unified policy network with spatial constraint reasoning to enable safe human following under uncertain and dynamic pedestrian environments.
\item We extend CrowdNav with static obstacles and conduct extensive evaluations with varying crowd densities, pedestrian behaviors, and environment layouts, and deploy on a real robot under the ROS 2 setup to validate real-world effectiveness.
\end{itemize}

\section{Related Work}
\subsection{Human-Following Robot}

Human-following robots stay near a target human while moving, forming the foundation for many human-robot interaction tasks~\cite{islam2019person, eirale2025human}. Existing research focuses on two aspects: perception and decision making. For perception, approaches range from transmitter-based localization~\cite{scheidemann2025obstacle} to vision-based tracking under partial occlusion~\cite{ye2023robot} and recent vision-language methods~\cite{wang2025trackvla}, though the latter can suffer from ambiguity when language cannot precisely describe the target or under out-of-distribution conditions. For decision making, optimization-based methods embed following objectives into explicit cost formulations~\cite{song2023safe, situ2025adap}, while learning-based approaches~\cite{kastner2022human, leisiazar2023mcts} enable fast inference via end-to-end policies. However, most are evaluated in in-distribution settings or with constrained motion patterns, lacking analysis of dense crowds where humans exhibit complex OOD behaviors. Although we develop a tracking system for the target human, our core focus is on explicitly managing the proximity-safety trade-off in dense crowds through constraint decomposition.

\subsection{Constrained Reinforcement Learning}
CRL has gained increasing attention as an alternative to reward engineering~\cite{kim2024not, lee2023evaluation} and in safe robot learning~\cite{yao2025towards, brunke2022safe, yao2024sonic}, as it allows the integration of constraints into RL agents during the learning process, ensuring that the agents follow these constraints. Several methods are commonly employed, including Constrained Policy Optimization (CPO)~\cite{achiam2017constrained}, Projection-Based Constrained Policy Optimization (PCPO)~\cite{yang2020projection}, and Lagrangian methods~\cite{ray2019benchmarking}. Among these, Lagrangian methods are particularly advantageous as they can be integrated with any existing RL algorithms and are relatively easy to implement. In this work, we build on PPO-Lagrangian~\cite{ji2024omnisafe, ray2019benchmarking} and decompose the human-following task into multiple independent cost constraints with behaviorally meaningful thresholds, rather than encoding all objectives into a single reward.

\section{Method}\label{methods}

\subsection{Problem Formulation}
We consider an environment with \( H \) humans and \( O \) static obstacles, indexed by \( h \) and \( o \), over an episode with horizon \( T \). 
One human is designated as the target \( h_{\text{target}} \), whom the robot must follow while maintaining a distance \( d_{\text{follow}} \leq d_{\text{valid}} \).
We formulate this task as a Constrained Markov Decision Process (CMDP)~\cite{altman2021constrained}. At each timestep \( t \), the agent observes a state \( S_t \) comprising the robot's current state, the positions and predicted trajectories of nearby humans from an upstream predictor, and local occupancy grid maps representing static obstacles, and generates an action \( A_t = (v_x, v_y) \) controlling the robot's velocity.
Executing the action returns a reward \( R_t \) and three distinct cost signals:
1) \textit{Following cost} \( C^{\text{F}}_t \), penalizing deviation from the target;
2) \textit{Human collision cost} \( C^{\text{H}}_t \), penalizing proximity to other humans;
and 3) \textit{Obstacle collision cost} \( C^{\text{O}}_t \), penalizing proximity to static obstacles.
This decomposition enables independent tuning of the robot's behavior for each aspect of the task.
Our objective is to learn an optimal policy $\pi(A_t \mid S_t)$ that maximizes cumulative reward while keeping human and obstacle safety costs below their respective limits $\delta^H$ and $\delta^O$, and maintaining the following cost at a desired level $\delta^F$, where each threshold carries direct behavioral meaning, such as tolerable intrusions to pedestrians or obstacles, or a desired following distance to the target.

\subsection{Method Overview}
\begin{figure}[!tbp]
\centering
\includegraphics[width=1.0\columnwidth]{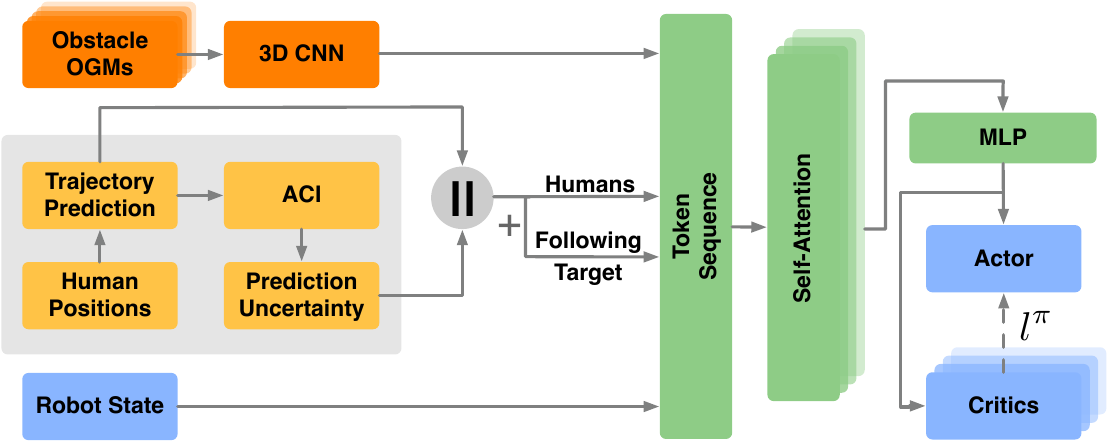}
\caption{An overview diagram of our method. Components related to 
static obstacles are highlighted in orange, those related to humans 
in yellow, the shared feature extraction modules in green, and those 
concerning the robot's physical state and decision making in blue. 
The detailed mechanism by which the critics generate $l^{\pi}$ is 
illustrated in Fig.~\ref{fig:critics}.}
\label{fig:overview}
\vspace{-0.4cm}
\end{figure}

\begin{figure}[!tbp]
	\centering
	\includegraphics[width=0.85\columnwidth]{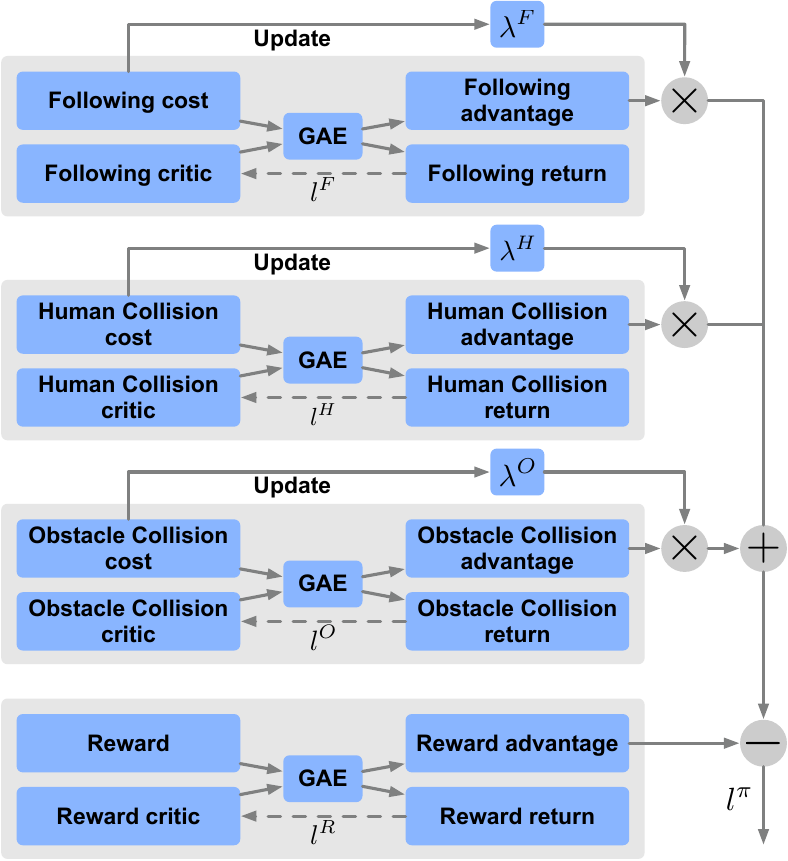}
    \vspace{-0.1cm}
	\caption{The interaction among the critics and the generation of the action loss. More details can be found in Section \ref{training}.
    \vspace{-0.4cm}
    }
	\label{fig:critics}
\end{figure}

An overview of our method is illustrated in Fig.~\ref{fig:overview}. 
At each timestep $t$, the state $S_t$ consists of the robot's physical state, local occupancy grid maps representing static obstacles, and the current positions, predicted trajectories, and quantified prediction uncertainties of nearby humans.
The state is processed by a unified policy network, where CNN-encoded occupancy features and human and robot tokens are combined into a sequence and processed by self-attention mechanisms~\cite{vaswani2017attention}, producing a feature representation that serves as input to the actor-critic framework.
We employ an actor-critic framework for policy learning, where the actor network is guided by multiple critics. As shown in Fig.~\ref{fig:critics}, the four critics include: a reward critic corresponding to the task reward, a following cost critic for 
maintaining proximity to the target human, an obstacle collision cost critic, and a human collision cost critic, where uncertainty estimates are further incorporated into the cost formulation.
Each critic applies Generalized Advantage Estimation (GAE)~\cite{schulman2016high} to compute advantages and returns. These critics jointly guide the actor's policy updates via PPO-Lagrangian~\cite{ji2024omnisafe, ray2019benchmarking}, where Lagrangian multipliers are dynamically adjusted to enforce the cost constraints on following and collision avoidance.

\subsection{Uncertainty-Aware Policy Network for Safe Following}\label{sec:network}
We design a Transformer-based policy network that reasons about complex social interactions among agents and spatial constraints from static obstacles for safe human following in dense crowds.
Spatial-temporal scene context is encoded using a 3D convolutional neural network (CNN) that extracts features from local occupancy grid maps (OGMs) over the most recent 5 time steps, leveraging the inductive bias of convolutions for capturing local spatial patterns in grid-structured data. 
For all detected humans, we construct uncertainty-aware feature tokens comprising the current position, a sequence of $K$ predicted future positions, and their quantified uncertainties:
\begin{equation}
\resizebox{0.88\hsize}{!}{$
\mathbf{h}_h(t) = [\mathbf{p}_h(t), \mathbf{p}_{h,1}(t), \ldots, 
\mathbf{p}_{h,K}(t);\hat\delta_{h,1}(t), \ldots, \hat\delta_{h,K}(t)],
$}
\end{equation}
where $\mathbf{p}_h(t)$ is the current position of the $h$-th human at time $t$, $\mathbf{p}_{h,k}(t) = (x_{h,k}(t)\,, y_{h,k}(t))$ denotes the $k$-th predicted position, and $\hat{\delta}_{h,k}(t)$ is the estimated prediction uncertainty for the $k$-th step. 
The uncertainty is computed by an adaptive conformal inference (ACI) module~\cite{lindemann2023safe, yao2025towards} that maintains online error bounds adapted to the actual prediction quality. At each timestep $t$, we compute the prediction error $\delta_{h,k}(t) = \| \mathbf{p}_h(t) - \mathbf{p}_{h,k}(t{-}k) \|_2$ 
for the $k$-th horizon of the $h$-th human, where $\mathbf{p}_h(t)$ is the observed position at time $t$ and $\mathbf{p}_{h,k}(t{-}k)$ is the $k$-step prediction calculated at time $t{-}k$. We run $M$ parallel estimators to estimate the error bound, 
updated as
\begin{equation}
\hat{\delta}^{(m)}_{h,k}(t{+}1) = \hat{\delta}^{(m)}_{h,k}(t) + 
\gamma^{(m)} \left( \mathbf{1}\!\left[\delta_{h,k}(t) > 
\hat{\delta}^{(m)}_{h,k}(t)\right] - \alpha \right),
\end{equation}
where $\gamma^{(m)}$ is the learning rate of the $m$-th estimator and $\alpha \in (0,1)$ is the target miscoverage rate. The final uncertainty estimate $\hat{\delta}_{h,k}(t)$ is selected from the $M$ estimators via adaptive weighted sampling based on their cumulative performance. These per-human, per-horizon estimates provide the policy with an adaptive measure of prediction reliability, and the same ACI bounds are also incorporated into the human safety cost design in Sec.~\ref{training}.

To combine information from all agents and scene elements, we implement early fusion~\cite{nayakanti2023wayformer} by embedding the robot state $\mathbf{r}(t)$, target information $\mathbf{tg}(t)$, extracted obstacle features $\mathbf{o}(t)$, and human features $\mathbf{h}(t)$ into a shared representation space:
$\mathbf{x}(t) = [\, \mathbf{r}(t),\, \mathbf{tg}(t),\, 
\mathbf{o}(t),\, \mathbf{h}(t)],$
where $\mathbf{tg}(t)=\mathbf{h}_{target}$, and $\mathbf{x}(t) \in \mathbb{R}^{N \times d}$, with $N$ representing the total number of tokens and $d$ the embedding dimension. 
We represent each entity, including the robot, the target, perceived obstacles, and all detected humans, as an individual feature embedding, allowing the self-attention mechanism to model all pairwise interactions across the full scene. In particular, obstacle-human and obstacle-robot relationships can be explicitly captured, whereas these interactions are lost when obstacle features are appended after the attention 
stage~\cite{liu2026height}. Human tokens are ordered by each human's distance to the robot, allowing the positional encoding to reflect proximity and enabling the attention mechanism to prioritize nearby agents.

We apply sinusoidal positional encoding~\cite{vaswani2017attention} and a binary mask for undetected humans:
$\mathbf{x}_{\text{pe}}(t) = \mathrm{PE}(\mathbf{x}(t)) 
\odot \mathbf{M}(t),$
where $\mathbf{M}(t) \in \{0,1\}^{N \times d}$ zeroes out tokens of undetected humans. 
The encoded sequence is processed by a Transformer encoder as $\mathbf{z}(t) = \mathcal{T}(\mathbf{x}_{\text{pe}}(t))$, and we extract the robot token output $\mathbf{z}_r(t) \in \mathbb{R}^d$ to produce the final representation $\mathbf{f}_r(t) = \mathrm{MLP}\!\left(\mathbf{z}_r(t)\right)$, which serves as input to our CRL policy module described in Sec.~\ref{training}.

\subsection{Task Decomposition for Explicit Behavior Control}\label{training}

Human following in dense crowds requires simultaneously minimizing collision risk and maintaining an appropriate following distance, two inherently conflicting objectives.
Encoding all these objectives into a single reward function makes it difficult to independently control each behavior, as reward weights implicitly determine the trade-off and cannot be directly mapped to interpretable behavioral outcomes~\cite{kim2024not}. 
We instead formulate the problem as a CMDP~\cite{altman2021constrained} that decouples task success from behavioral constraints, enabling explicit and independent control over each aspect:
\begin{equation}
\resizebox{0.90\hsize}{!}{$
\begin{gathered}
    \max_{\pi} \mathbb{E}_{\tau \sim \pi} \left[ \sum_{t=0}^{T} 
    R_t(S_t, A_t) \right] \; \text{s.t.} 
    \ \mathbb{E}_{\tau \sim \pi} \left[ \sum_{t=0}^{T} 
    C^{\text{F}}_t \right] = \delta^{\text{F}}, \\
    \mathbb{E}_{\tau \sim \pi} \left[ \sum_{t=0}^{T} 
    C^{\text{H}}_t \right] \leq \delta^{\text{H}}, \;
    \mathbb{E}_{\tau \sim \pi} \left[ \sum_{t=0}^{T} 
    C^{\text{O}}_t \right] \leq \delta^{\text{O}},
\end{gathered}
$}
\end{equation}
where $R_t$ represents the reward function, $C^{\text{F}}_t$, $C^{\text{H}}_t$, and $C^{\text{O}}_t$ denote the cost functions for following behavior, human safety, and obstacle safety, respectively, and $\delta^{\text{F}}$, $\delta^{\text{H}}$, $\delta^{\text{O}}$ are the corresponding cost thresholds. The safety costs use inequality constraints since safer behavior is always preferable, while the following cost uses an equality constraint, maintaining a desired distance rather than minimizing it to avoid both target loss and personal-space intrusion.

Our reward function provides sparse feedback only for the task outcomes:
\begin{equation}
R_t(S_t, A_t) = 
\begin{cases}
    R_{\text{success}}, & \text{if } S_t \in S_{\text{success}}, \\
    R_{\text{collision}}, & \text{if } S_t \in 
    S_{\text{collision}}, \\
    R_{\text{target\_lost}}, & \text{if } S_t \in 
    S_{\text{target\_lost}},
\end{cases}
\end{equation}
while all behavioral objectives are managed through the three cost constraints. Each cost function isolates a specific behavior, and its constraint threshold $\delta^i$ provides direct and interpretable control over the desired trade-off.

First, the following cost $C^{\text{F}}_t$ measures the robot's distance beyond the personal space threshold $d_{\text{personal}}$, encouraging the robot to maintain a moderate following distance rather than either lagging behind or crowding the target:
\begin{equation} \resizebox{0.90\hsize}{!}{$
    C^{\text{F}}_t = 
    \begin{cases}
        k_1 (d_{\text{follow}, t} - d_{\text{personal}}), 
        & \text{if } d_{\text{follow}, t} > d_{\text{personal}}, \\
        0, & \text{otherwise},
    \end{cases} $}
\end{equation}
where $d_{\text{follow}, t}$ is the Euclidean distance between the robot and the target at time $t$, $d_{\text{personal}}$ is the personal space threshold, and $k_1$ is the penalty coefficient.

Second, the human safety cost $C^{\text{H}}_t$ quantifies the risk of collision with other pedestrians by measuring intrusions into uncertainty-aware safety regions around each human. For each human $h$, we define a safety region around the current position $\mathbf{p}_h(t)$ with radius $r_{\text{ego}} + r_h + r_{\text{buf}}$, where $r_{\text{ego}}$ and $r_h$ are the radii of the robot and human, and $r_{\text{buf}}$ is a fixed buffer margin. To account for the inherent uncertainty in pedestrian motion prediction, we additionally define safety regions around each of the first $K'$ 
predicted positions $\mathbf{p}_{h,k}(t)$ with radius $r_{\text{ego}} + r_h + \hat{\delta}_{h,k}(t)$, where the ACI uncertainty bound $\hat{\delta}_{h,k}(t)$ from Sec.~\ref{sec:network} scales each safety region according to the local prediction 
confidence at that horizon, expanding protection when motion is uncertain and contracting it when predictions are reliable. The cost is proportional to the maximum intrusion depth across all humans and all safety regions at each timestep:
\begin{equation}
    C^{\text{H}}_t = k_2 \, d_{\mathrm{intru}, t},
\end{equation}
where $k_2$ is the penalty coefficient and $d_{\mathrm{intru},t}$ is the maximum intrusion distance. This formulation aims to ensure safety under unpredictable pedestrian behaviors by grounding spatial protection in quantified prediction uncertainty.

Third, the obstacle safety cost penalizes the robot for getting too close to static obstacles:
\begin{equation} \resizebox{0.88\hsize}{!}{$
C^{\text{O}}_t = \begin{cases} 
0, & \text{if } \min_o(d^{\text{surf}}_{o, t}) \geq 
d_{\text{safe\_o}}, \\ 
k_3(d_{\text{safe\_o}} - \min_o(d^{\text{surf}}_{o, t})), 
& \text{if } \min_o(d^{\text{surf}}_{o, t}) < d_{\text{safe\_o}},
\end{cases}
$}
\end{equation}
where $\min_o(d^{\text{surf}}_{o, t})$ is the minimum surface distance from the robot to all static obstacles at time $t$, $d_{\text{safe\_o}}$ is the safe distance threshold, and $k_3$ is the penalty coefficient.

We employ PPO-Lagrangian~\cite{ray2019benchmarking} to optimize a single unified policy under the three cost constraints, with four separate critics estimating the value functions for the reward and each cost independently:
\begin{equation} \resizebox{0.88\hsize}{!}{$
\begin{gathered}
    l^R_t = c_1 (V^R_{\theta_1}(S_t) - V_t^{\text{targ}, R})^2, 
    \quad
    l^{\text{F}}_t = c_2 (V^{\text{F}}_{\theta_2}(S_t) - 
    V_t^{\text{targ}, {\text{F}}})^2, \\
    l^{\text{H}}_t = c_3 (V^{\text{H}}_{\theta_3}(S_t) - 
    V_t^{\text{targ}, {\text{H}}})^2, \quad
    l^{\text{O}}_t = c_4 (V^{\text{O}}_{\theta_4}(S_t) - 
    V_t^{\text{targ}, {\text{O}}})^2,
\end{gathered} $}
\end{equation}
where $c_1$ through $c_4$ are loss weighting coefficients, $V^{(\cdot)}_{\theta_i}(S_t)$ are the predicted value functions, and $V_t^{\text{targ},(\cdot)}$ are the corresponding target values computed from collected rollouts.

We compute four distinct advantage estimates using GAE~\cite{schulman2016high}, including the reward advantage $\hat{A}^R_t$ and three cost advantages $\hat{A}^{\text{F}}_t$, 
$\hat{A}^{\text{H}}_t$, $\hat{A}^{\text{O}}_t$. These are combined into a single advantage signal:
\begin{equation}
    \hat{A}^{\prime}_t = \frac{
        \hat{A}^R_t
        - \lambda^{\text{F}} \hat{A}^{\text{F}}_t
        - \lambda^{\text{H}} \hat{A}^{\text{H}}_t
        - \lambda^{\text{O}} \hat{A}^{\text{O}}_t
    }{
        1 + \lambda^{\text{F}} + \lambda^{\text{H}} 
        + \lambda^{\text{O}}
    },
\end{equation}
where $\lambda^{\text{F}}$, $\lambda^{\text{H}}$, $\lambda^{\text{O}}$ are Lagrangian multipliers, and the denominator ensures the combined advantage has a consistent 
scale regardless of the number of active constraints. The actor is updated using the PPO clipping objective~\cite{schulman2017proximal}:
\begin{equation}
l^{\pi}_t = -\min\!\left( 
\rho_t \hat{A}'_t,\; 
\mathrm{clip}(\rho_t, 1{-}\epsilon, 1{+}\epsilon) 
\hat{A}'_t \right),
\end{equation}
where $\rho_t = \pi_\theta(A_t|S_t) / \pi_{\theta_{\text{old}}}(A_t|S_t)$ is the importance sampling ratio and $\epsilon$ is the clipping parameter. The Lagrangian multipliers are updated via dual gradient descent after each epoch:
\begin{equation}
\lambda^{i} \leftarrow \max\!\left(0,\; \lambda^{i} + 
\eta_\lambda \left( \hat{J}_{C^i}(\pi) - \delta^{i} 
\right)\right), \quad i \in \{\text{H}, \text{O}\},
\end{equation}
\begin{equation}
\lambda^{\text{F}} \leftarrow \lambda^{\text{F}} + 
\eta_\lambda \left( \hat{J}_{C^{\text{F}}}(\pi) - \delta^{\text{F}} 
\right),
\end{equation}
where $\eta_\lambda$ is the multiplier learning rate and $\hat{J}_{C^i}(\pi)$ is the empirical average cost. When the current cost exceeds its limit, the corresponding multiplier increases, amplifying the cost advantage's contribution to $\hat{A}'_t$ and steering the policy toward constraint satisfaction, and vice versa. This mechanism provides automatic and interpretable control over each behavioral constraint, directly linking the cost thresholds $\delta^i$ to the robot's safety-performance trade-off.
While multiple critics are employed during training, only the actor network is needed at inference, so the computational cost at deployment is identical to single-critic approaches.

\section{Experiments}

\subsection{Simulation Settings}
We construct cluttered indoor environments fully enclosed by walls, measuring up
to \SI{20}{m} in both dimensions, with randomly generated static obstacles of
varying sizes and locations to ensure scene-level diversity. The robot and 40
humans are placed in the scene, with the robot initialized at a random position
with a maximum speed of \SI{1.2}{m/s} and a target human sampled within
\SI{1.6}{m}. The remaining 39 humans are randomly positioned and controlled by
ORCA~\cite{van2011reciprocal}, with radii sampled between \SI{0.3}{m}--\SI{0.4}{m}
and maximum speeds between \SI{0.7}{m/s}--\SI{1.4}{m/s}. Once a human reaches
its goal, a new goal is assigned.

\subsection{Evaluation Metrics}
Our evaluation metrics include: 
1) \textit{Success Rate (SR)}: SR is defined as the ratio of successful following episodes to the total number of test episodes. 
2) \textit{Collision Rate (CR)}: CR is the ratio of episodes in which the robot collides with a human or an obstacle. 
3) \textit{Target Lost Rate (TLR)}: TLR is the ratio of episodes in which the distance between the robot and the target human $d_{\text{follow}}$ exceeds the valid threshold $d_{\text{valid}}$ at any time during the episode. 
4) \textit{Average Following Distance (AFD)}: AFD is the average distance between the robot and the target human throughout the entire episode. 

\begin{table}[!tbp]
    \centering
    \caption{In-Distribution Test Results}
    \footnotesize
    \renewcommand{\arraystretch}{1.1}
    \setlength{\tabcolsep}{1mm}
    \begin{tabular}{
        l
        c
        c
        c
        c
        c
        c
    }
        \toprule
        \multirow{2}{*}{\textbf{Methods}} & \multirow{2}{*}{\textbf{SR}$\uparrow$} 
        & \multicolumn{3}{c}{\textbf{CR}$\downarrow$} 
        & \multirow{2}{*}{\textbf{TLR}$\downarrow$} 
        & \multirow{2}{*}{\textbf{AFD}} \\
        \cmidrule(lr){3-5}
        & & \textbf{Overall} & \textbf{Human} & \textbf{Obstacle} \\
        \midrule 
        SG-HA* & 1.84\% & 81.52\% & 79.92\% & \textbf{1.60}\% & 16.64\% & 1.91 \\
        SG-ORCA & 17.68\% & 64.88\% & 27.84\% & 37.04\% & 17.44\% & 1.65 \\
        SG-MPC & 30.96\% & 42.72\% & 30.08\% & 12.64\% & 26.32\% & 2.64 \\
        OGM-HEIGHT & 52.32\% & 34.72\% & 21.44\% & 13.28\% & 12.96\% & 2.19 \\
        \midrule
        RL & 68.00\% & 24.08\% & 15.76\% & 8.32\% & 7.92\% & 2.24 \\
        RL+ACI & 71.60\% & 20.72\% & 12.80\% & 7.92\% & 7.68\% & 2.28 \\
        \midrule
        Ours & \textbf{78.08}\% & \textbf{16.16}\% & \textbf{10.72}\% & \textbf{5.44}\% & \textbf{5.76}\% & 2.35 \\
        \bottomrule
    \end{tabular}
    \vspace{-0.5cm}
    \label{in_distribution_result}
\end{table}

\subsection{Baselines and Ablation Models}
We compare our CRL-based framework against representative baselines. Following the common two-step paradigm of subgoal generation and downstream planning~\cite{zhang2021efficient, song2023safe, scheidemann2025obstacle}, we construct three subgoal-guided baselines using the subgoal generation strategy 
from~\cite{scheidemann2025obstacle}:
1) \textit{SG-HA*}: Hybrid A*~\cite{dolgov2010path}, a search-based planner with continuous motion primitives;
2) \textit{SG-MPC}: An optimization-based controller that incorporates following task objectives~\cite{song2023safe, situ2025adap} and an uncertainty-aware cost function for 
dynamic human and static obstacle avoidance~\cite{huang2025interaction};
3) \textit{SG-ORCA}: ORCA~\cite{van2011reciprocal}, a classic velocity-based collision avoidance algorithm. 
We also include the state-of-the-art RL-based method HEIGHT~\cite{liu2026height}:
4) \textit{OGM-HEIGHT}: HEIGHT with its point cloud input replaced by OGMs for simulation compatibility, retrained to convergence with a following distance penalty in place of the original navigation reward, where the OGMs preserve the same local obstacle geometry as the point cloud input for a fair comparison. 
To validate the contributions of our uncertainty integration 
and cost decomposition, we include two ablation models:
5) \textit{RL}: Our policy network trained with standard RL and the same reward function as OGM-HEIGHT, without uncertainty estimates in the observations; and
6) \textit{RL+ACI}: Our full policy network trained with standard RL, where the three behavioral objectives are incorporated as weighted penalty terms into a single scalar reward, with the human safety penalty remaining uncertainty-aware using the same ACI-derived bounds as our method. 
For both RL+ACI and our method, we report the best result among all tuned configurations (see Table~\ref{tuning_comparison}).

\subsection{Implementation Details}

We train on an NVIDIA RTX 4090 GPU with a batch size of 480 and a clip parameter of 0.02 for PPO-Lagrangian. The valid following threshold $d_\text{valid} = \SI{5.0}{m}$ also serves as the perception range, the personal distance is $d_\text{personal} = \SI{1.0}{m}$ beyond which following costs are incurred, and the obstacle safety distance is $d_\text{safe\_o} = \SI{0.50}{m}$. The policy network uses a Transformer encoder with 4 layers and 8 attention heads, and we evaluate 1250 samples across 5 random seeds. Trajectory predictions use a constant velocity (CV) model~\cite{scholler2020constant}, which is simple yet sufficient for our setting: the online adaptation of ACI does not assume a highly accurate predictor, but instead expands uncertainty bounds as prediction errors grow, maintaining conservative safety regions under the non-linear pedestrian behaviors in our OOD scenarios.

\begin{table}[!tbp]
    \centering
    \caption{Effect of Constraint Tuning vs.\ Reward Weight Tuning}
    \renewcommand{\arraystretch}{1.1}
    \setlength{\tabcolsep}{1mm}
    \resizebox{\columnwidth}{!}{
    \begin{tabular}{
        l
        l
        c
        c
        c
        c
        c
    }
        \toprule
        \multirow{2}{*}{\textbf{Intent}} & \multirow{2}{*}{\textbf{Method}} 
        & \multirow{2}{*}{\textbf{SR}$\uparrow$} 
        & \multicolumn{2}{c}{\textbf{CR}$\downarrow$} 
        & \multirow{2}{*}{\textbf{TLR}$\downarrow$} 
        & \multirow{2}{*}{\textbf{AFD}} \\
        \cmidrule(lr){4-5}
        & & & \textbf{Overall} & \textbf{Human} \\
        \midrule
        \multirow{2}{*}{Safety}
        & Ours ($\delta_F$=4.0, $\delta_H$=3.2) & 71.68\% & 18.24\% & \textbf{8.80}\% & 10.08\% & 2.54 \\
        & RL+ACI ($w_H \times 2$) & 71.60\% & 20.72\% & 12.80\% & 7.68\% & 2.28 \\
        \midrule
        \multirow{2}{*}{Following}
        & Ours ($\delta_F$=3.2, $\delta_H$=4.0) & 74.40\% & 22.40\% & 14.24\% & \textbf{3.20}\% & 2.23 \\
        & RL+ACI ($w_F \times 2$) & 63.92\% & 27.36\% & 15.44\% & 8.72\% & 2.26 \\
        \midrule
        \multirow{2}{*}{Balanced}
        & Ours ($\delta_F$=3.6, $\delta_H$=3.6) & \textbf{78.08}\% & \textbf{16.16}\% & 10.72\% & 5.76\% & 2.35 \\
        & RL+ACI ($w_F$=$w_H$=1) & 70.80\% & 21.12\% & 12.88\% & 8.08\% & 2.26 \\
        \bottomrule
    \end{tabular}
    }
    \vspace{-0.5cm}
    \label{tuning_comparison}
\end{table}

\begin{table*}[!tbp]
    \centering
    \caption{Out-of-Distribution Test Results}
    \vspace{-0.2cm}
    \fontsize{6.5}{5.5}\selectfont
    \renewcommand{\arraystretch}{1.0}
    \setlength{\tabcolsep}{1mm}
    \resizebox{0.94\textwidth}{!}{
        \begin{tabular}{
            m{2.5cm}<{\raggedright} 
            m{1.8cm}<{\raggedright} 
            m{1.3cm}<{\centering} 
            m{1.3cm}<{\centering} 
            m{1.3cm}<{\centering} 
            m{1.3cm}<{\centering} 
            m{1.3cm}<{\centering} 
            m{1.3cm}<{\centering} 
        }
            \toprule
            \multirow{2}{*}{\textbf{Environments}} 
            & \multirow{2}{*}{\textbf{Methods}} 
            & \multirow{2}{*}{\textbf{SR}$\uparrow$} 
            & \multicolumn{3}{c}{\textbf{CR}$\downarrow$} 
            & \multirow{2}{*}{\textbf{TLR}$\downarrow$} 
            & \multirow{2}{*}{\textbf{AFD}} \\
            \cmidrule(lr){4-6}
            & & & \textbf{Overall} & \textbf{w/ Humans} & \textbf{w/ Obstacles} \\
            \midrule 
            \multirow{7.6}{*}{Corridor}
            & SG-HA* & 2.64\% & 88.64\% & 88.56\% & \textbf{0.08}\% & 8.72\% & 1.46 \\
            & SG-ORCA & 50.64\% & 47.20\% & 25.28\% & 21.92\% & 2.16\% & 1.48 \\
            & SG-MPC & 48.88\% & 39.20\% & 37.04\% & 2.16\% & 11.92\% & 1.84 \\
            & OGM-HEIGHT & 56.24\% & 37.60\% & 28.96\% & 8.64\% & 6.16\% & 1.50 \\
            & RL & 77.52\% & 21.84\% & 21.04\% & 0.80\% & 0.64\% & 1.50 \\
            & RL+ACI & 82.96\% & 16.72\% & 15.12\% & 1.60\% & \textbf{0.32}\% & 1.53 \\
            & Ours & \textbf{89.76}\% & \textbf{8.64}\% & \textbf{8.48}\% & 0.16\% & 1.60\% & 1.78 \\
            \midrule
            \multirow{7.6}{*}{15\% Rushing Humans}
            & SG-HA* & 0.16\% & 93.68\% & 92.72\% & \textbf{0.96}\% & 6.16\% & 1.93 \\
            & SG-ORCA & 12.48\% & 81.76\% & 37.28\% & 44.48\% & 5.76\% & 1.39 \\
            & SG-MPC & 15.76\% & 72.48\% & 57.84\% & 14.64\% & 11.76\% & 2.66 \\
            & OGM-HEIGHT & 47.04\% & 38.08\% & 26.56\% & 11.52\% & 14.88\% & 2.22 \\
            & RL & 61.60\% & 35.36\% & 27.36\% & 8.00\% & \textbf{3.04}\% & 2.15 \\
            & RL+ACI & 68.48\% & 28.48\% & 23.04\% & 5.44\% & \textbf{3.04}\% & 2.18 \\
            & Ours & \textbf{70.56}\% & \textbf{22.72}\% & \textbf{19.36}\% & 3.36\% & 6.72\% & 2.23 \\
            \midrule
            \multirow{7.6}{*}{SF Pedestrian Model}
            & SG-HA* & 0.72\% & 84.88\% & 84.08\% & \textbf{0.80}\% & 14.40\% & 1.74 \\
            & SG-ORCA & 22.24\% & 68.80\% & 14.40\% & 54.40\% & 8.96\% & 1.53 \\
            & SG-MPC & 29.60\% & 45.92\% & 31.20\% & 14.72\% & 24.48\% & 2.43 \\
            & OGM-HEIGHT & 35.84\% & 50.40\% & 22.88\% & 27.52\% & 13.76\% & 2.16 \\
            & RL & 56.96\% & 40.96\% & 21.28\% & 19.68\% & 2.08\% & 2.15 \\
            & RL+ACI & 60.64\% & 34.48\% & 21.68\% & 12.80\% & 4.88\% & 2.19 \\
            & Ours & \textbf{64.32}\% & \textbf{33.68}\% & \textbf{20.08}\% & 13.60\% & \textbf{2.00}\% & 2.21 \\
            \midrule
            \multirow{7.6}{*}{Groups}
            & SG-HA* & 0.08\% & 92.64\% & 92.64\% & 0.00\% & 7.28\% & 1.83 \\
            & SG-ORCA & 27.68\% & 46.48\% & 46.48\% & 0.00\% & 25.84\% & 1.86 \\
            & SG-MPC & 26.88\% & 64.08\% & 64.08\% & 0.00\% & 9.04\% & 2.43 \\
            & OGM-HEIGHT & 36.32\% & 53.28\% & 53.28\% & 0.00\% & 10.40\% & 2.09 \\
            & RL & 49.36\% & 46.64\% & 46.64\% & 0.00\% & \textbf{4.00}\% & 1.87 \\
            & RL+ACI & 56.00\% & 39.20\% & 39.20\% & 0.00\% & 4.80\% & 2.17 \\
            & Ours & \textbf{59.68}\% & \textbf{32.64}\% & \textbf{32.64}\% & 0.00\% & 7.68\% & 2.14 \\
            \bottomrule
        \end{tabular}
    }
    \vspace{-0.2cm}
    \label{ood_result}
\end{table*}

\subsection{In-Distribution Results}
The test results under the same setting as the training environment are shown in Table~\ref{in_distribution_result}.
Traditional subgoal-guided methods (SG-HA*, SG-ORCA, SG-MPC) exhibit substantially lower 
SR and higher CR and TLR compared to RL-based methods, indicating limited ability to maintain following behavior in the presence of dense, dynamic pedestrians. 
Among RL-based methods, our approach significantly outperforms OGM-HEIGHT, achieving a 25.76\% higher SR while reducing overall CR by 18.56\%, human CR by 10.72\%, and obstacle CR to the lowest 5.44\% among learning-based methods. The ablation results further show that \textit{RL+ACI} improves over \textit{RL} in human CR, confirming that uncertainty-aware observations enable more cautious behavior around unpredictable pedestrians, and our full model achieves the best overall performance, outperforming the two ablation models by 10.08\% and 6.48\% in SR while reducing human CR by 5.04\% and 2.08\%, obstacle CR by 2.88\% and 2.48\%, and TLR by 2.16\% and 1.92\%. These results validate the effectiveness of our cost decomposition in achieving a good balance between proximity and safety.

\begin{figure*}[!tbp]
    \centering
    \includegraphics[width=\textwidth]{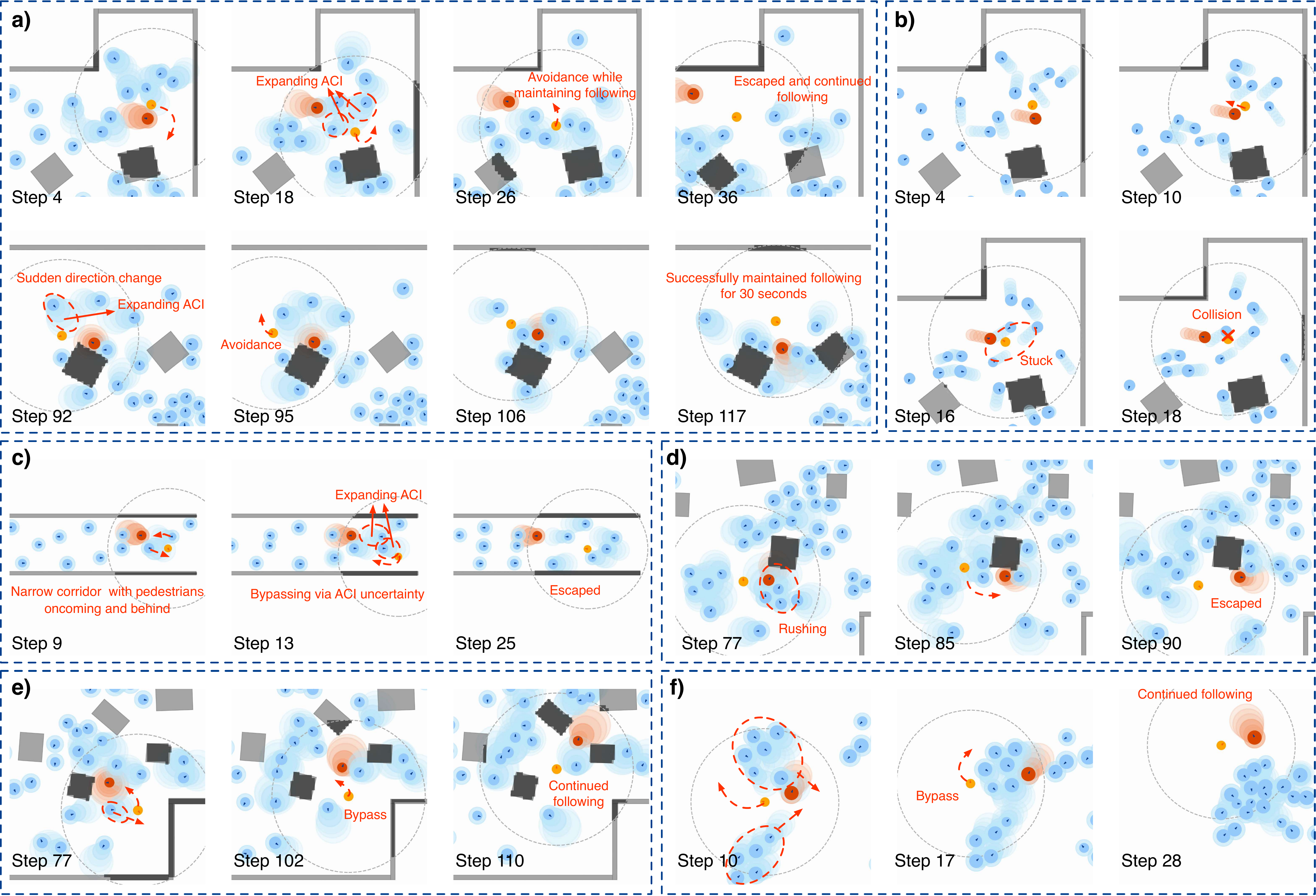}
    \vspace{-0.5cm}
    \caption{Visualization of test results. Regular pedestrians are shown in blue, the target pedestrian is represented in red, and the robot is shown in orange. Light blue circles surrounding human agents represent spatial safety buffers derived from our uncertainty quantification method. 
    (a) Ours successfully completes the human-following task in an in-distribution testing environment. 
    (b) OGM-HEIGHT fails to complete the same episode; the light blue circles here indicate trajectory predictions rather than safety buffers. 
    (c) Ours in an OOD narrow corridor environment. 
    (d) Ours in an OOD environment with rushing humans. 
    (e) Ours in an OOD environment using the SF pedestrian model. (f) Ours in an OOD environment with pedestrian groups.}
    \label{fig:QualitativeAnalysis}
    \vspace{-0.2cm}
\end{figure*}

We visualize the behaviors of Ours and OGM-HEIGHT in the same episode in Fig.~\ref{fig:QualitativeAnalysis}(a) and Fig.~\ref{fig:QualitativeAnalysis}(b). OGM-HEIGHT moves directly toward the target but becomes trapped as pedestrians begin moving, ultimately colliding. Our method instead responds to expanding uncertainty areas by steering toward less crowded regions while maintaining the general direction and proximity toward the target at steps 4 and 18. At step 92, when a pedestrian suddenly changes direction, our robot adjusts its trajectory based on the updated uncertainty bounds, successfully avoiding collision while maintaining target proximity. This demonstrates that cost decomposition and uncertainty-aware cost formulation enable the robot to effectively balance safety and following under dynamic and unpredictable pedestrian behaviors.

\subsection{Cost Limits vs. Reward Weights for Behavioral Control}

To validate that explicit cost constraints provide more interpretable and predictable
behavioral control than reward-based penalty encoding, we compare cost limit tuning
against reward weight tuning under three behavioral intents in
Table~\ref{tuning_comparison}. In both cases, the same policy network architecture
and ACI-derived uncertainty bounds are used. For RL+ACI, $w_F$, $w_H$, and $w_O$
denote the penalty weights for the following distance, human safety, and obstacle safety
in the reward function, respectively. Since the primary conflict in pedestrian crowd following lies between proximity to the target and safety among dynamic pedestrians,
we fix $\delta_O = 1.2$ and $w_O$ across all configurations and focus the comparison
on the following and human safety objectives.

We configure three constraint profiles that shift the system's behavioral
priority: safety-conservative ($\delta_F$=4.0, $\delta_H$=3.2), balanced
($\delta_F$=3.6, $\delta_H$=3.6), and aggressive-following ($\delta_F$=3.2,
$\delta_H$=4.0). 
As the profile shifts from safety-conservative to aggressive-following, AFD
decreases from 2.54 to 2.23 and TLR drops from 10.08\% to 3.20\%, while human
CR rises from 8.80\% to 14.24\%. Tightening $\delta_F$ thus produces closer
following, whereas tightening $\delta_H$ yields safer behavior, each achieved
at the expense of the other and reflecting predictable control over both
objectives. The balanced profile sits between these extremes and achieves the
highest SR of 78.08\% and the lowest overall CR of 16.16\%, offering the best
compromise. These results show that each cost threshold maps directly and
predictably onto its intended behavioral aspect, allowing the desired
proximity-safety trade-off to be specified explicitly.

In contrast, reward weight tuning produces inconsistent and counterintuitive
outcomes. Doubling $w_H$ yields a negligible 0.08\% reduction in human CR, from 12.88\% to 12.80\%, failing to reflect the intended safety priority. More
strikingly, doubling $w_F$ leaves AFD completely unchanged at 2.26 while
simultaneously raising TLR from 8.08\% to 8.72\% and human CR from 12.88\%
to 15.44\%, making both following and safety worse. These results confirm that cost limit tuning provides a direct and verifiable mapping from designer specification to policy behavior, which reward weight tuning fundamentally cannot offer.

\subsection{Out-of-Distribution Results}

We evaluate the performance of our policy and baselines under four representative OOD scenarios. 

\textit{1) OOD Scenarios in the Corridor Environment:} We construct a \SI{26}{m} $\times$ \SI{4.5}{m} corridor with 20 randomly initialized pedestrians, and the flow naturally splits into a \textit{bidirectional pedestrian flow}. 
As shown in Table~\ref{ood_result}, most baseline and ablation methods exhibit markedly higher human-collision rates in this narrow, crowded environment, whereas our method maintains a low human collision rate of 8.48\%. It reduces the human-collision rate by 20.48\% compared with OGM-HEIGHT and by 12.56\% and 6.64\% compared with the two ablation models, and also achieves higher success rates, exceeding OGM-HEIGHT by 33.52\% and the two ablations by 12.24\% and 6.80\%. A representative episode is visualized in Fig.~\ref{fig:QualitativeAnalysis}(c): at step 9, pedestrians advance from the opposite direction while those behind accelerate, yet our policy identifies a feasible escape path and maintains an appropriate following distance to the target.

\textit{2) OOD Scenarios Mixed with 15\% Rushing Humans:} In this setting, 15\% of pedestrians are assigned faster walking speeds, with a minimum of \SI{1.4}{m/s} and a maximum of \SI{1.7}{m/s}, which corresponds to the typical speed of brisk walking. From the results in Table~\ref{ood_result}, we observe that the SR of all methods drops. Nevertheless, our method continues to achieve the best overall performance. We also visualize an example in Fig.~\ref{fig:QualitativeAnalysis}(d), showing that even under the influence of rushing pedestrians, our approach can follow the target effectively and safely.

\textit{3) OOD Scenarios with SF Pedestrian Model:} 
In this setting, all pedestrian agents follow the social force (SF) model. By modeling social forces between agents, humans' speeds change more markedly, and in denser scenarios, pedestrians tend to move closer together and create cohesive units. As expected, most methods experience a decline in success rate, which indicates limited adaptability to different behavioral models. We visualize a representative test case in Fig.~\ref{fig:QualitativeAnalysis}(e), showing that our model is still able to avoid collisions and sustain robust following of the target human.

\textit{4) OOD Scenarios with Group Dynamics:} In this setting, pedestrians move in groups in an open environment, which occupy larger spatial areas and require wider detours, so it also leads to a drop in performance across most methods. As visualized in Fig.~\ref{fig:QualitativeAnalysis}(f), our robot can navigate around clustered groups while still successfully keeping up with the target human.

Through extensive experiments under different conditions, we find that the policy is most likely to break in scenarios with dense and fast-moving crowds, as well as unpredictable pedestrian behaviors such as sudden turns and closely aligned group movements.

\subsection{Real-World Robot Experiments}
We deploy the policy trained in simulation on a ROSMASTER X3 robot equipped with Mecanum wheels running ROS 2 after basic clipping and smoothing to validate real-world feasibility. Perception relies on a 2D RPLIDAR-A1 LiDAR, with human detection via the pre-trained DR-SPAAM model~\cite{jia2020dr}, tracking via SORT~\cite{bewley2016simple}, and trajectory prediction via a CV predictor~\cite{scholler2020constant}, over which our ACI module maintains online uncertainty bounds. Across 10 test trajectories in varied settings, the robot successfully completes 7, following the target human while navigating among moving pedestrians and static obstacles. The three failures stem from upstream perception limitations, such as missed detections of dynamic obstacles and inaccurate static boundary estimations, as well as the target human exceeding the robot's maximum speed, while the navigation policy itself remains robust throughout all runs. Further demonstrations are available on our project page.

\section{Conclusion}

This paper addresses the proximity-safety conflict in human-following 
navigation through a multi-constraint RL framework that decomposes 
behavioral objectives into independent cost constraints with 
behaviorally meaningful thresholds, enabling explicit and tunable 
trade-off control. We integrate prediction uncertainty into 
the policy network and cost formulation to improve safety under 
unpredictable pedestrian conditions. Extensive experiments across in-distribution and out-of-distribution scenarios demonstrate 
improvements over baselines, and direct comparison confirms that cost 
threshold tuning provides more predictable behavioral control than 
reward weight tuning. Real-robot deployment validates 
real-world feasibility. Future work will explore adaptive policies that incorporate individual human preferences.



\bibliographystyle{IEEEtran}
\bibliography{References}

\end{document}